\documentclass[conference]{IEEEtran}
\IEEEoverridecommandlockouts
\usepackage{amsmath,amssymb,amsfonts}
\usepackage{algorithmic}
\usepackage{graphicx}
\usepackage{textcomp}

\usepackage{color}	
\usepackage[dvipsnames,table,xcdraw]{xcolor}
\usepackage{listings} 
\usepackage{biblatex} 
\def\BibTeX{{\rm B\kern-.05em{\sc i\kern-.025em b}\kern-.08em
    T\kern-.1667em\lower.7ex\hbox{E}\kern-.125emX}}
\begin{document}

\renewcommand{\*}{\cdot}
\newcommand{\tabitem}{~~\llap{\textbullet}~~}
\definecolor{backcolour}{rgb}{0.85,0.85,0.85}
    \lstset{
    language=C,
    backgroundcolor=\color{backcolour},
    commentstyle=\color{PineGreen},
    keywordstyle=\color{WildStrawberry},
    numberstyle=\color{black},
    stringstyle=\color{MidnightBlue},
    basicstyle=\ttfamily\footnotesize,
    frame=bt,
    captionpos=b,
    breakatwhitespace=false,         
    breaklines=true,                 
    keepspaces=true,                 
    numbers=left,       
    numbersep=5pt,                  
    showspaces=false,                
    showstringspaces=false,
    showtabs=false,                  
    tabsize=2           
    }

\title{A Continual and Incremental Learning Approach for TinyML On-device Training Using Dataset Distillation and Model Size Adaption
}
\author{\IEEEauthorblockN{1\textsuperscript{st} Marcus Rüb \& 2\textsuperscript{nd} Philipp Tuchel}
\IEEEauthorblockA{\textit{Software Solutions / Artificial intelligence} \\
\textit{Hahn-Schickard}\\
Villingen-Schwenningen, Germany \\
Name.Surname@hahn-schickard.de}
\and
\IEEEauthorblockN{3\textsuperscript{rd} Axel Sikora}
\IEEEauthorblockA{\textit{Institute of Reliable Embedded Systems}\\
\textit{and Communication Electronics (ivESK)} \\
\textit{Offenburg University of Applied Sciences}\\
Offenburg, Germany \\
Axel.Sikora@hs-offenburg.de}
\and
\IEEEauthorblockN{4\textsuperscript{th} Daniel Mueller-Gritschneder}
\IEEEauthorblockA{\textit{Electronic Design Automation} \\
\textit{Technical University of Munich}\\
Munich, Germany \\
Daniel.Mueller@tum.de}

}

\maketitle

\begingroup
\let\clearpage\relax

\begin{abstract}
A new algorithm for incremental learning in the context of Tiny Machine learning (TinyML) is presented, which is optimized for low-performance and energy efficient embedded devices. 
TinyML is an emerging field that deploys machine learning models on resource-constrained devices such as microcontrollers, enabling intelligent applications like voice recognition, anomaly detection, predictive maintenance, and sensor data processing in environments where traditional machine learning models are not feasible. The algorithm solve the challenge of catastrophic forgetting through the use of knowledge distillation to create a small, distilled dataset. The novelty of the method is that the size of the model can be adjusted dynamically, so that the complexity of the model can be adapted to the requirements of the task. This offers a solution for incremental learning in resource-constrained environments, where both model size and computational efficiency are critical factors. Results show that the proposed algorithm offers a promising approach for TinyML incremental learning on embedded devices. The algorithm was tested on five datasets including: CIFAR10, MNIST, CORE50, HAR, Speech Commands. The findings indicated that, despite using only 43\% of Floating Point Operations (FLOPs) compared to a larger fixed model, the algorithm experienced a negligible accuracy loss of just 1\%. In addition, the presented method is memory efficient. While state-of-the-art incremental learning is usually very memory intensive, the method requires only 1\% of the original data set.
\end{abstract}


\section{Introduction}

Machine learning has conventionally been carried out on centralized infrastructures, with models trained on substantial datasets before deployment on end-user devices. However, the rise of mobile and IoT devices has accelerated the shift towards on-device training \cite{Sharma.2021}, where models are trained directly on devices like smartphones and microcontrollers. This method offers benefits such as enhanced privacy, decreased latency, and better energy efficiency due to local data processing\cite{Ren.15.03.2021}. Despite its advantages, on-device training on resource-constrained devices such as Micro-Controller Units (MCUs) poses challenges, especially with memory capacity and computing power \cite{Rub.2022}.

One significant challenge for on-device training is the constraint of storing large datasets, making traditional post-training techniques unviable. This has led to the exploration of incremental or continuous learning strategies. Continuous learning allows models to acquire knowledge across time without forgetting previous learnings, unlike incremental learning which focuses on new class acquisition. However, a significant hurdle of continuous learning is the phenomenon of catastrophic forgetting, causing rapid degradation of a model's performance on previous tasks when trained on new ones. Many current methods to address this are memory-intensive.

To address these challenges, this paper makes the following contributions:
\begin{itemize}
    \item Proposes a novel approach for scalable and efficient incremental learning on embedded devices.
    \item Demonstrates continuous on-device training with adaptive model sizing according to the data.
    \item Introduces an algorithm that starts with a compact model, ensuring low memory and computational resource usage, with potential incremental expansions.
    \item Validates the approach on 5 diverse datasets: CIFAR10, MNIST, CORe50, HAR, Speech Commands, achieving models with minimal accuracy loss despite drastic reductions in memory and computational resources.
\end{itemize}

\section{related work}

We compare ourselves with three popular techniques in incremental learning that do not heavily rely on extra memory. First, CWR (CopyWeights and Re-init, see experiments in \cite{lomonaco2017core50}) essentially skips certain layers and fixes weights for other layers in order to maintain information about past inputs. A similar approach, called EWC (Elastic Weight Consolidation, \cite{kirkpatrick2017overcoming}) is an algorithm that allows for continuous learning by constraining important parameters in neural networks to stay close to their old values. The constraint is implemented as a quadratic penalty which puts weight on the performance of previous tasks. The third one is LwF (Learning without forgetting, \cite{li2017learning}) and works by using a distillation loss function to transfer knowledge from the original model to the new model. The distillation loss encourages the new model to output similar probabilities to the original model for the data that was present in the original training set. This helps the new model to learn the new task while minimizing interference with the knowledge it has already acquired. 

In \cite{parisi2018lifelong}, they compared those three methods with the cumulative approach (our baseline) and the naive approach (catastrophic forgetting). This is done in the same scenario as we do (CORe50, NIC scenario). The cumulative approach, CWR, LwF and EWC archived 64.13\%, 29.56\%, 28.94\% and 28.l31\% accuracy, respectively. The baseline got 19.39\%. In their experiments, they use a slightly larger model, i.e. it makes sense to compare the relative accuracies (also, they use 128x128 resolution, while we use only 64x64). However, our method goes one step further and can change with the complexity of the new data and, for example, increase in size. as a result, we only use as much computing power as necessary.

There are further techniques that use extra memory in order to maintain the accuracy from previous steps. For instance, in \cite{lopes2017data}, a student neural network uses a larger teacher network's precomputed activation records. These records are used to reconstruct the original dataset without accessing the original data. The reconstructed dataset is then used as training input for the student network. Similarly, in DGR (deep Generative Replay, see \cite{shin2017continual}) the generated input samples were paired with “hard targets” provided by the main model. A deep generative model (“generator”) and a task-solving model (“solver”) work together. With these two models,  training data for previous tasks can be sampled and interleaved with those for a new task. Finally, the method in \cite{parisi2018lifelong} avoids catastrophic forgetting by using growing networks. In contrast to those approaches, our technique is simpler and more memory efficient. 

There are also different ways to construct a distillation dataset. One can distill the dataset by matching the gradients of the whole dataset and the distilled one. This method was proposed in \cite{zhao2020dataset}. An improvement was done by implementing DSA (Differentiable Siamese Augmentation) which allows an application of augmentation techniques that can be backtracked in the neural network in order to obtain gradients (\cite{zhao2021dataset}). Finally, in \cite{zhao2023dataset}, the dataset is distilled by matching the distribution of the final layer of the whole dataset with a small synthesized dataset. In our experiments we use the last technique (\cite{zhao2023dataset}) as it performs best in our setting. It is worth noting that this procedure can also be replicated with different distillation methods and thus can also be adapted for future techniques. 

\begin{figure*}[!hbt] 
	\centering
	\includegraphics[width=0.7\textwidth]{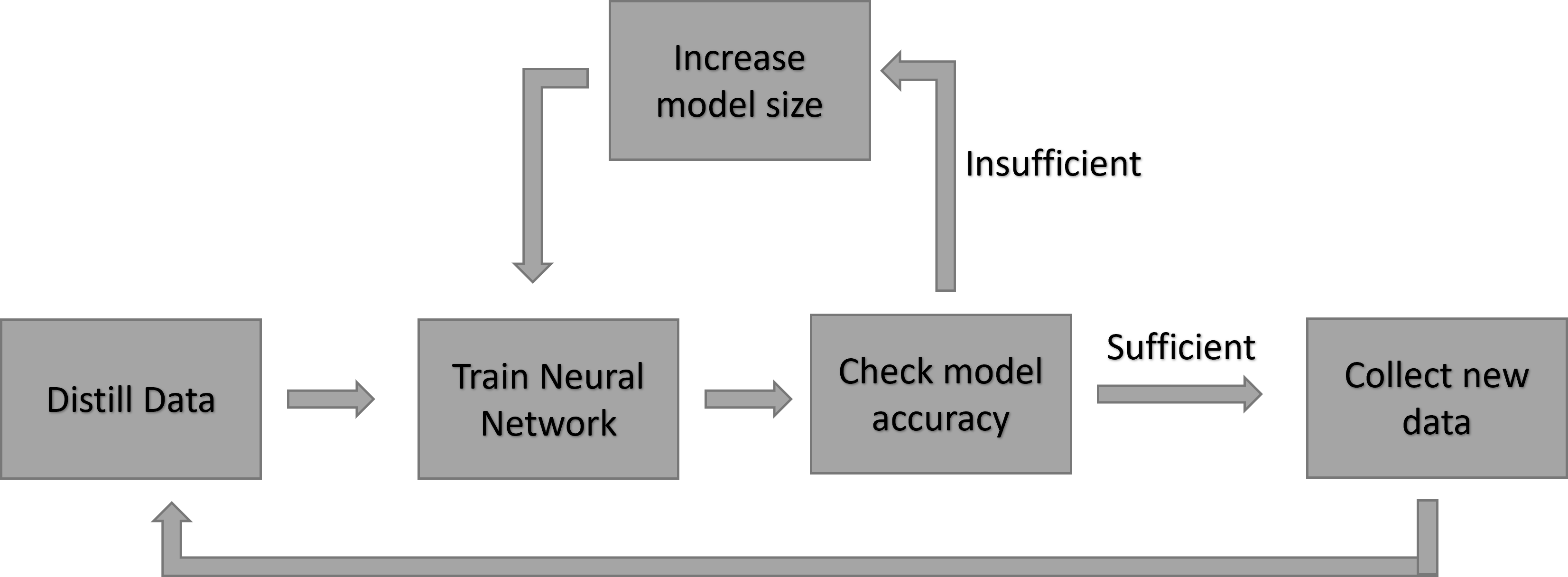} 
	\caption{The presented method  to solve the problem consists of the following steps: 1. Compression of existing data
2. Training of the neural network
3.Adjusting the size of the network as needed.
4. Deployment of the trained network and collection of new data
5. Adding new data to improve performance}
	\label{process}
\end{figure*}

\section{Proposed Method}
\subsection{Problem Statement}
In this section, we present our general pipeline for all the experiments. We consider a continuous learning or incremental learning scenario in the following sense: Consider a discrete number of $N$ timestamps. At timestamp $t$, only the current dataset $\mathcal{X}_t$ is available and can be used for training of a neural network. The accuracy of the current and all previously seen data (which is not available) is equally essential. Assume the existence of a test set $\mathcal{T}_t$ corresponding to the data of each step $t$. After $n$ steps, the best possible accuracy on the entire test $\{\mathcal{T}_t\}_{t=1}^n$ and small computational cost and memory are of interest. We assume the network to run on resource-constrained devices with small capability. In order to record the increase of performance over time, we evaluate the model at each step on the entire test set $\{\mathcal{X}_t\}_{t=1}^n$. 

\subsection{Method}
The aforementioned problem of limited memory and computational resources on embedded devices has posed significant challenges in the research and development of on-device training. Our solution to this problem is to use compressed datasets and smaller neural network architectures specifically designed for embedded devices. In this context, we present a multi-step method to address this challenge. \\
Our method encapsulates a cyclic process subdivided into five stages, which are sequentially executed. The process is shown in Fig.~\ref{process}.

1. Data Distillation: The foremost step involves distilling the existing data. This process involves carefully selecting the most crucial features and data points relevant to our application, ultimately creating a condensed dataset that requires significantly less storage space.

2. Neural Network Training: Post data distillation, we utilize this compressed dataset to train our neural network. This reduced dataset size not only accelerates the training process but also conserves computational resources, making it an efficient approach for embedded devices.

   3. Model Accuracy Validation: Following the training phase, we evaluate our model's accuracy. This step ensures that our model is providing outputs that align as closely as possible to the desired results.

    4. Model Enlargement (If Necessary): If the model's accuracy does not meet the predefined standards—i.e., it performs poorly— we then move to an iterative process of model enlargement. This entails increasing the complexity of our neural network and revisiting step 2 to re-train this more sophisticated model.

    5. Data Recording and Iteration: If the model exhibits satisfactory performance, we record the next dataset and the aforementioned process commences anew. Thus, this system ensures our model's consistent growth and improvement, adapting to new data and conditions over time.

The procedure encapsulates a balance between resource utilization and model performance, making it a potentially effective solution for continuous learning scenarios on resource-constrained devices.

\subsubsection{Distill Data}

Data distillation is a very active research area. Even though our results and accuracies highly depend on the data distillation method, the overall workflow can be applied to almost all of them. In this paper, we use the distribution matching method for images by \cite{DBLP:journals/corr/abs-2110-04181} for illustration, which will be shortly explained in the following. We start by initializing a certain number of images per class (IPC) (e.g. 1 or 10) which we consider as the distilled data. Now, for $K$ steps, we randomly initialize a new convolutional model and measure the difference from the feature space in the last convolutional layer in the network of real data to the feature space of the distilled images. More precisely, we calculate the loss
$$
\mathcal{L} = \sum_{c=1}^C ||\frac{1}{|\mathcal{X}|}\sum_{x\in\mathcal{X}[c]}\psi(x) -\frac{1}{|\mathcal{X}^\prime|}\sum_{x\prime\in\mathcal{X}^\prime[c]}\psi(x^\prime) ||
$$
where, $C$ is the number of classes, $\mathcal{X}$ and $\mathcal{X}^\prime$ is the real- and distilled datasets respectively, $(\cdot)[c]$ the subset only containing class $c$ and $\psi$ is map into a feature space depending on the neural network. Finally, we apply a gradient descent update w.r.t. the distilled data, i.e. $\mathcal{X}^\prime_{i+1} = \mathcal{X}^\prime_{i} - \eta\nabla_{\mathcal{X}^\prime_{i}}\mathcal{L_i}$ in step $i+1$, where $\eta$ is a learning rate. Consequently, $\mathcal{X}^\prime$ converges to a similar feature space as the real data, which should result in a similar training performance. We also apply Differentiable Siamese Augmentation (DSA) (omitted in the above notation) which is an augmentation technique suitable for backpropagation (\cite{zhao2021dataset}). For a more detailed analysis, see \cite{DBLP:journals/corr/abs-2110-04181}.

Compression not only saves storage space but also reduces the computational resources needed to train the neural network.

\subsubsection{Neural Network Training}

Following the data distillation process, the next stage involves the utilization of this condensed dataset to train our neural network. This phase is essential, as it is here that the distilled data is ingested into the neural network and the weights of the network are adjusted in an iterative manner. The ultimate goal of this step is to calibrate the output of the network to approximate the target results as accurately as possible.

Due to the reduced size of the distilled dataset, the training phase is faster and requires fewer computational resources. This advantage makes our method particularly apt for application in embedded devices with limited computational capabilities.

The training process of a neural network follows a generally iterative methodology and can be modeled using a loss function. This function is utilized to quantify the degree of divergence between the predicted output of the model and the true output. Let's denote the cost function as $\mathcal{C}$. The network training is then performed using an optimization process that aims to find the model parameters that minimize the cost function.

The cost function $\mathcal{C}$ is often expressed as an average over the cost functions $\mathcal{C}_x$ for each individual training sample $x$. If we denote the total number of training examples as $n$, the cost function $\mathcal{C}$ can be expressed as:

$$
\mathcal{C} = \frac{1}{n} \sum_x \mathcal{C}_x
$$

The learning process of the neural network, often referred to as backpropagation, aims to adjust the weights and biases of the model to reduce the cost function. This process uses the gradient descent optimization algorithm, which involves the iterative adjustment of the parameters in the direction of the steepest descent of the cost function. This procedure can be represented as follows:

$$
w_{k+1} = w_k - \eta \nabla_w \mathcal{C}
$$
$$
b_{k+1} = b_k - \eta \nabla_b \mathcal{C}
$$

In these equations, $w_k$ and $b_k$ denote the weights and biases of the model at the $k^{th}$ iteration, respectively, $\nabla_w \mathcal{C}$ and $\nabla_b \mathcal{C}$ represent the gradients of the cost function with respect to the weights and biases, and $\eta$ is the learning rate, which determines the step size in the gradient descent algorithm.

The overall process ensures that our network, which never actually sees the real data but rather uses the distilled dataset $\{\mathcal{X}_t^\prime\}_{i=1}^t$, achieves an efficient balance between resource usage and training performance. This makes our approach particularly beneficial in scenarios where computational resources are limited, such as embedded devices.

\subsubsection{Model Accuracy Validation}

Once the training of the neural network has concluded, we need to verify the model's accuracy. The purpose of this step is to ascertain that our trained model is delivering outputs that are as close to the target results as possible.

For this purpose, we use a validation set $\{\mathcal{T}_t\}_{t=1}^n$ which is separate from our training set $\{\mathcal{X}_t^\prime\}_{i=1}^t$. It is crucial to evaluate model accuracy on unseen data to prevent overfitting to the training set and to ensure the model generalizes well to new data. 

Model accuracy is typically measured using an accuracy score, which is defined as the proportion of correct predictions out of the total number of predictions. If we denote the number of correct predictions as $C_{\text{correct}}$ and the total number of predictions as $N_{\text{total}}$, the accuracy score $A$ can be calculated as follows:

$$
A = \frac{C_{\text{correct}}}{N_{\text{total}}}
$$

It is important to note that accuracy is only one of many performance metrics that can be used to evaluate the model. Depending on the task, other metrics such as precision, recall, F1 score, area under the ROC curve (AUC-ROC), mean squared error (MSE), etc., may be more appropriate. The choice of metric depends on the specific needs and constraints of your task and application.

If the model's accuracy score doesn't meet the predefined standards (i.e., it performs poorly), we move to the next step: Model Enlargement.

In our approach, at every step $t$, we evaluate the model on the entire test set $\{\mathcal{T}_t\}_{t=1}^n$. The goal is to maximize accuracy on the entire test set over all timestamps while ensuring small computational cost and memory usage. 

By repeating this process for each $t$ in $N$ timestamps, we ensure the model performs well not only on the current data but also on all previously seen data, even though that data is not available at the current timestamp. This approach allows us to efficiently tackle the continuous learning scenario within the constraints of limited resources.

\subsubsection{Model Enlargement (If Necessary)}

The model enlargement stage is an optional phase initiated when the model's accuracy doesn't meet predefined standards or benchmarks—meaning the model is underfitting the data. Underfitting signifies that the model fails to capture the underlying patterns in the data adequately, thereby delivering poor performance during the accuracy validation step.

Model enlargement involves adding complexity to the neural network architecture, such as increasing the number of layers or neurons per layer, or changing the types of layers (e.g., adding convolutional layers for image tasks or LSTM layers for sequential data).

Let's denote our original model's architecture as $\mathcal{M}$ and its accuracy score as $A$. If $A$ doesn't meet the predefined accuracy standard, $A_{\text{standard}}$, we proceed with the enlargement of the model, creating a new model architecture, $\mathcal{M}'$. This adjustment can be represented as follows:

\[
\mathcal{M}' = 
\begin{cases} 
\mathcal{M} + \Delta\mathcal{M} & \text{if } A < A_{\text{standard}} \\
\mathcal{M} & \text{otherwise}
\end{cases}
\]

where $\Delta\mathcal{M}$ represents the increase in the model's complexity. This increase can be in the form of additional layers, more neurons per layer, inclusion of different types of layers, and so forth.

This more sophisticated model, $\mathcal{M}'$, is then retrained using the distilled data $\{\mathcal{X}_t^\prime\}_{i=1}^t$. The increased complexity allows the model to learn more intricate patterns within the data, ideally improving its accuracy score. After the training phase, the model is once again validated using the validation dataset $\{\mathcal{T}_t\}_{t=1}^n$.

This iterative process of enlargement and re-training continues until the model's accuracy meets or surpasses the predefined standards. However, care must be taken not to over-complicate the model, as this could lead to overfitting, where the model learns the training data too well and performs poorly on unseen data.

By adaptively adjusting the complexity of the model, this method ensures that the model remains appropriate for the data complexity at each step. Thus, this method is an efficient solution for incremental learning scenarios on resource-constrained devices, as it facilitates performance optimization without unnecessary resource consumption.

\subsubsection{Data Recording and Iteration}

The final step in this method is data recording and iteration. After the model has been trained and validated, and possibly enlarged, the process moves forward with the next dataset. This step involves recording or acquiring new data, which is then subjected to the complete sequence of steps described earlier.

Suppose $\mathcal{X}_{t+1}$ represents the new dataset obtained at the $(t+1)$-th timestamp, the cycle begins anew with this data. The recorded data is first distilled using the data distillation method described in earlier sections, creating a new distilled dataset, $\mathcal{X}_{t+1}^\prime$.

Following this, the distilled data is used to train the current neural network model. The model accuracy is then evaluated, and the model enlargement step is initiated if necessary. This cyclical process ensures consistent improvement and adaptation of the model to new data and conditions over time.

Summarizes, the process is as follows:

1. Data distillation:

   $$
   \mathcal{X}_{t+1}^\prime = \text{Distill}(\mathcal{X}_{t+1})
   $$

2. Neural network training:

   $$
   \mathcal{M}_{t+1} = \text{Train}(\mathcal{M}_t, \mathcal{X}_{t+1}^\prime)
   $$

3. Model accuracy validation:

   $$
   A_{t+1} = \text{Validate}(\mathcal{M}_{t+1}, \{\mathcal{T}_{t+1}\}_{t=1}^{n})
   $$

4. Model enlargement (if necessary):

   $$
   \mathcal{M}'_{t+1} = 
   \begin{cases} 
   \mathcal{M}_{t+1} + \Delta\mathcal{M}_{t+1} & \text{if } A_{t+1} < A_{\text{standard}} \\
   \mathcal{M}_{t+1} & \text{otherwise}
   \end{cases}
   $$

5. Data recording and iteration: $\mathcal{X}_{t+2}$ is recorded and the cycle continues.

By repeating this cyclic process for each timestamp $t$, this methodology ensures the neural network's consistent growth and adaptability to new data and conditions over time. It maintains a balance between resource utilization and model performance, making it an effective solution for continuous learning scenarios on resource-constrained devices.

This method has significant advantages over classical incremental- or continuous learning methods. We need much less computational resources since the network is only trained on very small datasets that lead to fast convergence. Furthermore, we can do not need to save the model during the entire training time, since it can be recovered by the distilled data quickly any time. This also provides the possibility to adapt the network complexity to the data complexity. In many scenarios we find it sufficient to use a less complex model in the beginning, since data complexity is small. As the number of steps increase, we switch to a more complex model, which is not possible for many other methods.

\section{Experiments}
In this section, we delve into the specifics of our experiments, shedding light on the chosen datasets and the experimental setup.

\subsection{Experimental Setup}
In all experiments, we use a convolutional neural network which was recently used in different scenarios involving distilled data (see e.g. \cite{zhao2021dataset}). Networks with depths of 2,3 and 4 blocks of convolutional layers are considered, each containing 8, 64, and 128 kernels, respectively (called ConvNetD2, ConvNetD3 and ConvNetD4). After each convolutional layer, a group normalization layer, ReLU activation function and a $2\times 2$ average pooling with stride $2$ follow. In order to model different complexities, we choose suitable value $D\in\{2,3,4\}$. This architecture is commonly used for data distillation benchmarks. Even though the data distillation process depends on the model architecture, we find little differences in performance w.r.t. the used model in the distillation, i.e. a ConvNetD2 performs similarly on distilled data generated on ConvNetD2 and on distilled data generated on ConvNetD4. Therefore, we always distill the data w.r.t. the biggest model ConvNetD4. This also allows the transition from a smaller to a larger model during incremental learning. In training, we apply DSA in the distillation process as well as a learning rate scheduler and stochastic gradient descent equipped with a momentum of $0.9$ and a weight decay of $5e-4$. Another reason to choose the relatively small networks ConvNetD2, ConvNetD3 and ConvNetD4 to simulate realistic TinyML scenarios at the expense of absolute accuracy for the larger datasets. In our standard setup, we use a learning rate of $0.01$, the Adam Optimizer and a batch size of $256$.
Each experiment consists of several steps in which the available dataset changes. We compare four scenarios. For the proposed method, we assume access to the currently available subsets at each step, as well as to the distilled images of all previously seen subsets. For the training with distilled data, two different cases are of interest. One consists of training only on the biggest ConvNetD4 model, which is also the one used for distillation. This is compared to the proposed adaptive model size scenario in which the network size increases with the number of steps, starting with the smallest ConvNetD2 network and, if necessary, upgrading to bigger networks as the training progresses. The needed flops for one epoch are compared for each training. For the baseline, only the data subset for the current step is available. This is the \emph{catastrophic forgetting scenario}. We also compare our method to the fully trained (cumulative) scenario, where all current and past subsets are completely available for training, which is of course the ideal case. After each step,  we test the model on the entire test set, if nothing else is mentioned. The exact results can be seen in the table I-V below.

\subsubsection{MNIST}
We use the MNIST dataset for handwritten digits to simulate a continuous learning scenario for changing data distribution in $10$ steps. The first step is standard training. In each other steps we modify the classes by rotating the images clockwise by a random angle: in step $i \in \{2,\dots ,10\}$ each training and test image gets rotated by an angle (in degree) $$\alpha_{train} \sim \text{U}([20(i-2),20(i-1)]), \quad \alpha_{test} \sim \text{U}([0,20(i-1)])$$
respectively, where $\text{U}$ is the uniform distribution.
This method causes some difficulties. For instance, the numbers $6$ and $9$ can not be distinguished for angles close to $180^{\circ}$. We distill 10 images per class from these randomly rotated training images. We omit DSA in this experiment, since it also contains rotation as an augmentation technique. In total, this is about $1.7\%$ of the data compared to the entire training data set.

\subsubsection{CORe50}
We preform an incremental learning scenario for the CORe50 dataset on object level classification (50 classes) and the NIC-scenario (new instances and new classes in every step). This is a rather difficult but highly realistic incremental learning setup, since new objects as well as new backgrounds are added each step. The first step contains of $10$ classes in one session. Each other step provides 5 classes of one session, resulting in 79 overall learning steps ($3$ sessions are used for testing). All images are rescaled to a size of $64$x$64$. We generate one distilled image for one class in one session leading to $400$ distilled images in total which is about $0.3\%$ is the entire dataset. Our method preserves over $78\%$ test accuracy of the fully (cumulative) trained network and prevents catastrophic forgetting. This also outperforms other popular incremental learning methods for CORe50 like CWR which maintains about $50\%$ and does not allow adaptive model size (see \cite{lomonaco2017core50}) Best results are archived by training a new network in every step only on the distilled images. There are at most 400 distilled images in each step, hence the training is highly efficient. The mixed training with current real and previous distilled images show more fluctuations and slightly worse results.

\subsubsection{Speech Commands}
In this experiment, we performed standard class incremental learning for the audio dataset Speech Commands with $35$ steps. After converting the audio data into mel frequency and a spectrogram, we passed it as image data with size $51x81$ into the same convolutional network as in the prior experiments. We distilled 20 images per class, which results in about $0.8\%$ of the size of the entire set.

\subsubsection{HAR}
The Human Activity Dataset contains sensor data which we again converted into mel spectrograms and passed it as images of size $26x31$ pixels into the convolutional network. The class incremental learning consists of $8$ steps. In this experiment, 10 IPC are used which is about $4.68\%$ of the data.

\subsubsection{CIFAR10}
Finally, we consider the classical CIFAR10 dataset in the class incremental learning scenario with $10$ steps. We distill 50 IPC, leading to $1\%$ of the size of the original dataset.

\section{Results and Discussion}
Within the results section, we provide a comprehensive visual analysis through Figure \ref{Result}, which illustrates the performance metrics of various models in relation to their computational complexity, measured in FLOPs.

\subsection{Explanation of Terms}
\begin{itemize}
    \item \textbf{Baseline}: Represents the model's performance when trained on the entire dataset without incremental learning. This serves as a reference point for comparing the efficiency and effectiveness of our incremental learning approach.
    \item \textbf{Largest Model}: This refers to a static, large model trained using the incremental learning approach but without adapting the model size.
    \item \textbf{Adaptive Model}: A dynamic model that adjusts its size during training, in line with the complexity of the task at each incremental learning step.
    \item \textbf{Catastrophic Forgetting}: A scenario where the model is trained only on the current data subset, without access to previous data, leading to a rapid decline in performance on previously learned tasks.
\end{itemize}

The outcomes are measured in terms of the end accuracy (accuracy achieved at the final stage), the average accuracy over time (across all steps), and the computation resources required (expressed as a percentage of floating point operations, or FLOPs).

\begin{figure*}[!hbt] 
	\centering
	\includegraphics[width=1\textwidth]{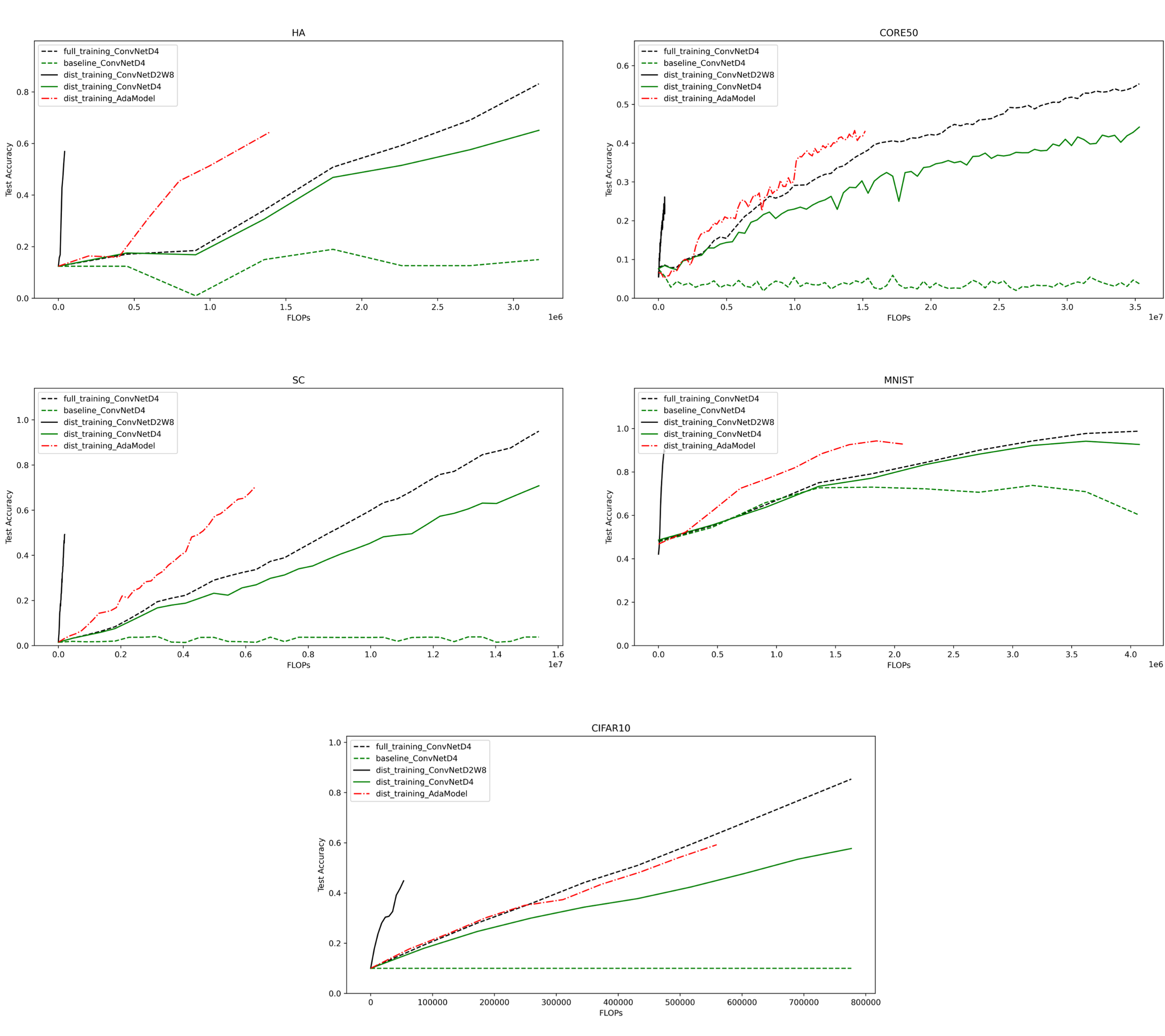} 
	
\caption{Comparison of models based on their performance plotted against the number of FLOPs. 
}

	\label{Result}
\end{figure*}

\subsection{CIFAR10}
As depicted in Table 1, our baseline model achieved an end accuracy of 85.4\%, averaging 47.8\% accuracy over all steps and requiring 100\% of the FLOPs. The largest model showed a decrease in performance with an end accuracy of 57.7\% and an average accuracy of 35.6\%, while still requiring the same amount of computational resources. Notably, the adaptive model demonstrated similar performance to the largest model but at a significantly lower computational cost, achieving an end accuracy of 58.2\% and an average accuracy of 35.9\% using only 43\% of the FLOPs. The catastrophic forgetting scenario, where only the data subset for the current step is available, resulted in an end and average accuracy of 10\%, indicating a considerable decline in model performance.

\begin{table}[h]
\resizebox{\columnwidth}{!}{%
\begin{tabular}{lllll}
                        & \textbf{End Accuracy} & \textbf{Average Accuracy} & \textbf{Needed FLOPs} \\
\textbf{Baseline}                & 85.4 \%        & 47.8 \%                      & 100 \%       \\
\textbf{Largest model}           & 57.7 \%        & 35.6 \%                      & 100 \%        \\
\textbf{Adaptive Model}          & 58.2 \%        & 35.9 \%                      & 43\%       \\
\textbf{Catastrophic Forgotting} & 10 \%        & 10 \%                      & 100 \%               
\end{tabular}%
}
\caption{Results for CIFAR10 Dataset: Comparing end accuracy, average accuracy over incremental learning steps, and computational resource requirements (FLOPs) for different models. }
\end{table}

\subsection{CORe50}
As can be seen in Table 2, the baseline model had an end accuracy of 55.3\% and an average accuracy of 36.5\% over time. The largest and adaptive models exhibited slightly lower performances, but the adaptive model required only half the computational resources (49\% of FLOPs). The catastrophic forgetting model fared the worst, with an end accuracy of 3.7\% and an average accuracy of 3.6\%.

\begin{table}[h]
\resizebox{\columnwidth}{!}{%
\begin{tabular}{lllll}
                        & \textbf{End Accuracy} & \textbf{Average Accuracy} & \textbf{Needed FLOPs} \\
\textbf{Baseline}                & 55.3 \%        & 36.5 \%                      & 100 \%                     \\
\textbf{Largest model}          & 44.2 \%        & 28.9 \%                      & 100 \%                     \\
\textbf{Adaptive Model}          & 43.2 \%        & 26.1 \%                      & 49\%                       \\
\textbf{Catastrophic Forgotting} & 3.7 \%        & 3.6 \%                      & 100 \%                      
\end{tabular}%
}
\caption{Results for CORe Dataset: Comparing end accuracy, average accuracy over incremental learning steps, and computational resource requirements (FLOPs) for different models. }
\end{table}

\subsection{Speech Commands}
In the Speech Commands experiment (Table 3), the baseline model attained a high end accuracy of 95\% and averaged 45.6\% over time. The largest and adaptive models reached around 70\% in terms of end accuracy, with the adaptive model achieving this using only 36\% of the FLOPs. Once again, the catastrophic forgetting model performed poorly, with an end accuracy of 3.8\% and an average accuracy of 2.9\%.

\begin{table}[h]

\resizebox{\columnwidth}{!}{%
\begin{tabular}{lllll}
                        & \textbf{End Accuracy} & \textbf{Average Accuracy} & \textbf{Needed FLOPs} \\
\textbf{Baseline}                & 95 \%        & 45.6 \%                      & 100 \%                       \\
\textbf{Largest model}            & 70.8 \%        & 35 \%                      & 100 \%                       \\
\textbf{Adaptive Model}          & 70.4 \%        & 33.7 \%                      & 36 \%                      \\
\textbf{Catastrophic Forgotting} & 3.8 \%        & 2.9 \%                      & 100 \%                      
\end{tabular}%
}
\caption{Results for Speechcommands Dataset: Comparing end accuracy, average accuracy over incremental learning steps, and computational resource requirements (FLOPs) for different models. }
\end{table}

\subsection{MNIST}
Table 4 illustrates the MNIST results. The baseline model achieved an impressive end accuracy of 98.8\% and an average accuracy of 78.8\%. The largest and adaptive models demonstrated comparable performances in the 92\% range for end accuracy, with the adaptive model achieving this using only 36\% of the FLOPs. Interestingly, the catastrophic forgetting model in this case was not as disastrous as in the other scenarios, reaching an end accuracy of 60.1\% and averaging 66.2\% over time.

\begin{table}[h]
\resizebox{\columnwidth}{!}{%
\begin{tabular}{lllll}
                        & \textbf{End Accuracy} & \textbf{Average Accuracy} & \textbf{Needed FLOPs} \\
\textbf{Baseline}                & 98.8 \%        & 78.8 \%                      & 100 \%                     \\
\textbf{Largest model}            & 92.7 \%        & 76.9 \%                      & 100 \%                     \\
\textbf{Adaptive Model}          & 92.8 \%        & 76.1 \%                      & 36\%                       \\
\textbf{Catastrophic Forgotting} & 60.1 \%        & 66.2 \%                      & 100 \%                    
\end{tabular}%
}
\caption{Results for MNIST Dataset: Comparing end accuracy, average accuracy over incremental learning steps, and computational resource requirements (FLOPs) for different models. }
\end{table}

\subsection{HAR}
As shown in Table 5, for the HAR dataset, the baseline model achieved an end accuracy of 83.1\% and averaged 43.1\% over time. The largest and adaptive models exhibited end accuracies in the mid-60\% range, with the adaptive model again reducing computational resources by nearly two-thirds. In this scenario, the catastrophic forgetting model resulted in a significant drop in performance, with an end accuracy of just 15\% and an average accuracy of 12.5\%.

\begin{table}[h]

\resizebox{\columnwidth}{!}{%
\begin{tabular}{lllll}
                        & \textbf{End Accuracy} & \textbf{Average Accuracy} & \textbf{Needed FLOPs} \\
\textbf{Baseline}                & 83.1 \%        & 43.1 \%                      & 100 \%                     \\
\textbf{Largest model}            & 65.1 \%        & 37.3 \%                      & 100 \%                     \\
\textbf{Adaptive Model}          & 64.4 \%        & 36.9 \%                      & 36\%                       \\
\textbf{Catastrophic Forgetting} & 15 \%        & 12.5 \%                      & 100 \%                      
\end{tabular}%
}
\caption{Results for HAR Dataset: Comparing end accuracy, average accuracy over incremental learning steps, and computational resource requirements (FLOPs) for different models.}
\end{table}

\subsection{Discussion}
In this section, we delve into the findings obtained from our experimentation on five diverse datasets. These results elucidate the successful implementation and benefits of the incremental learning approach proposed in this study, mainly focusing on memory efficiency and computational savings.

\subsubsection{Incremental Learning and Memory Efficiency}

Our experiments demonstrated the successful application of the incremental learning technique, especially in the context of on-device training where memory limitations are often a hindrance. The use of distilled datasets proved beneficial in combating the challenge of catastrophic forgetting typically associated with continuous learning. Remarkably, the memory overhead required for this technique was a mere 1\% of the total size of the original dataset. This reveals the potential of our proposed method in situations where device memory is a significant concern.

\subsubsection{Efficiencies of the Adaptive Model Size}

A unique aspect of our method is the adaptive expansion of our models, which helps to alleviate the demands on memory and computational resources. By incrementally increasing the model size as per the task requirements, we could avoid deploying an overly large model from the outset.

When examining the FLOPs required across the training process, our method demonstrated considerable savings. In the Speech Commands and HAR datasets, we only utilized 36\% of the FLOPs required by a fixed model. For the CIFAR10 dataset, this requirement dropped to 43\%, and even for the computationally heavy CORe50 dataset, our method needed only 49\% of the FLOPs. These reductions directly correlate with energy and time savings, making our technique particularly valuable for on-device training where these resources are scarce.

Further, our results showed minimal, if any, sacrifice in accuracy despite these significant savings. On average, the CIFAR10 dataset experienced a slight 0.3\% improvement in accuracy over the entire training process, while the CORe50 dataset recorded a marginal average loss of 2.8\% in accuracy. These outcomes suggest that the proposed method offers an effective balance between resource utilization and model performance.

In summary, our proposed method not only handles the inherent constraints of on-device training but also showcases efficient resource utilization while preserving model accuracy.

\section{Conclusions and Future Research}

Our research demonstrates that our incremental learning method enables efficient memory and computation, tested across five diverse datasets. We achieved minimal accuracy loss (averaging at 0.4\%) while necessitating a mere 1\% memory overhead relative to the complete dataset. Remarkably, the adaptive model size occasionally demanded just 36\% of the FLOPs throughout training compared to a static model size.

However, the method's prowess could waver based on dataset complexity and variability. Its dynamic model sizing, while resource-efficient, may not guarantee consistent peak performance. 

Future research can address these limitations and expand our method's horizons: Enhance Distillation Techniques, Optimize Model Adaptation, Diverse Task Exploration, Adaptation to Real-time Data Streams, Combine Efficiency Methods, Robustness Against Dataset Shifts

\endgroup

\printbibliography 


\end{document}